\documentclass[journal]{IEEEtran}

\ifCLASSINFOpdf
\else
\fi

\usepackage{graphicx}

\begin{document}

\title{Real-time emotion recognition for gaming using deep convolutional network features}

\author{S\'ebastien~Ouellet
\thanks{S\'ebastien Ouellet is with the DIRO, Universit\'e de Montr\'eal, Qu\'ebec, Canada, e-mail: sebouel@gmail.com.}}

\maketitle

\begin{abstract}
The goal of the present study is to explore the application of deep convolutional network features to emotion recognition. Results indicate that they perform similarly to recently published models at a best recognition rate of 94.4\%, and do so with a single still image rather than a video stream. An implementation of an affective feedback game is also described, where a classifier using these features tracks the facial expressions of a player in real-time.
\end{abstract}


\IEEEpeerreviewmaketitle

\section{Introduction}
\IEEEPARstart{D}{eep} convolutional neural networks have gained much popularity in recent years for vision-related applications since they were shown to achieve some of the highest accuracies in image classification tasks \cite{krizhevsky2012imagenet,ciresan2012multi}. Features extracted from these networks trained on classifying objects have also been applied to other tasks successfully with no further training \cite{donahue2013decaf}, such as style classification for photographs and paintings \cite{karayev2013recognizing}. These findings point to the potential of using such features for a generic visual system. The goal of this paper is to study whether these features can also perform well on an emotion recognition task.

\subsection{Related works}
The Extended Cohn-Kanade Dataset (CK+) is one of the most recent dataset compiled for emotion recognition and has a large number of participants as compared to other datasets commonly used before 2010 \cite{lucey2010extended}. In order to provide baseline performances, they implemented a multiclass SVM with Active Appearance Model features \cite{cootes2001active}. The average across emotions, found from the confusion matrix, shows a 83.3\% accuracy, with no blatant weakness for any emotion. Since its publication, many other researchers now evaluate their models on that dataset.

Khan et al. \cite{khan2012human} proposed a model based on human behavior data. They studied human participants through an eye-tracking experiment. By recording their gaze when shown images of facial expressions, they determined the salient regions of a human face for the individual emotions expressed in the CK+ dataset. Their model then extracts features from the regions of interest identified from the eye-tracking experiment and classifies them with a SVM. The features were computed as pyramid histograms of gradients (PHOG) \cite{bosch2007representing}. They report performances slightly above 95\% on the CK+ dataset, which is comparable to approaches published previously.    

The current state-of-the-art approach\footnote{To my knowledge} is a part-based model developed to detect the intensity of facial action units applied to emotions \cite{jeni2013continuous}, which are generally described in the literature as combinations of action units \cite{lucey2010extended,ekman1978facial}. Keypoints are extracted from faces and features are computed as a hand-designed sparse representation of patches around those keypoints. They also report performances on two different sets of the CK+ dataset: the onset and the apex of the emotion. The onset is limited to the first six frames of the sequence, which makes the task harder as the facial expressions are more subtle. The performance reported is 86\% for the onset, whereas the performance for the apex, i.e. the last few frames, is as high as 99\%.

As seen in the previously described papers, most approaches developed to solve emotion recognition use customized features for short sequences of facial expressions, and therefore they require particular efforts and might not be generalizable to other vision-related tasks. The interest of this paper lies not in developing a state-of-the-art system for that task, but in investigating whether features from an object recognition task can be transfered to achieve adequate results without further training, which would indicate a good generalizability. 

\section{Methods}
In the following sections, the system developed to classify emotions in still images and video streams will be described, as well as how its performance is evaluated. The implemementation is written in Python 2.7 under Ubuntu 13.10, and is reproducible with freely available software\footnote{The source code is hosted on Github at https://github.com/Zebreu/ConvolutionalEmotion. You can contact me at sebouel@gmail.com for details}.

\subsection{Deep convolutional neural network features}
To extract features from each image, a convolutional network model is used that was trained on 1.2 million images from ImageNet in the Large Scale Visual Recognition Challenge 2012, as described in \cite{krizhevsky2012imagenet}. A Python implementation of this model is distributed by Donahue et al. \cite{donahue2013decaf}\footnote{Available at https://github.com/UCB-ICSI-Vision-Group/decaf-release/} and integrated in the system currently described. All parameters were downloaded\footnote{Available at http://www.eecs.berkeley.edu/\textasciitilde jiayq/decaf\_pretrained/} to avoid retraining it. An important consideration is the difference in the images between the ImageNet Challenge and an emotion recognition task. Whereas the ImageNet Challenge contains images showing a very wide variety of objects, the differences across images in datasets such as CK+ are between individuals and their facial expressions. Especially in this case, no human faces are part of the ImageNet Challenge 2012, and it is therefore an object unknown to the network.

The model consists of seven layers (plus one logistic regression layer, which is of importance only for the ImageNet recognition task), five of which are convolutional, with the two remaining ones fully-connected. The output of every layer is accessible, and features of interest here are extracted from layers five (just before the image information goes through any fully-connected layer) and layer six. They respectively are of dimension 9126 and 4096.

\begin{figure}[!t]
\centering
\includegraphics[width=\linewidth]{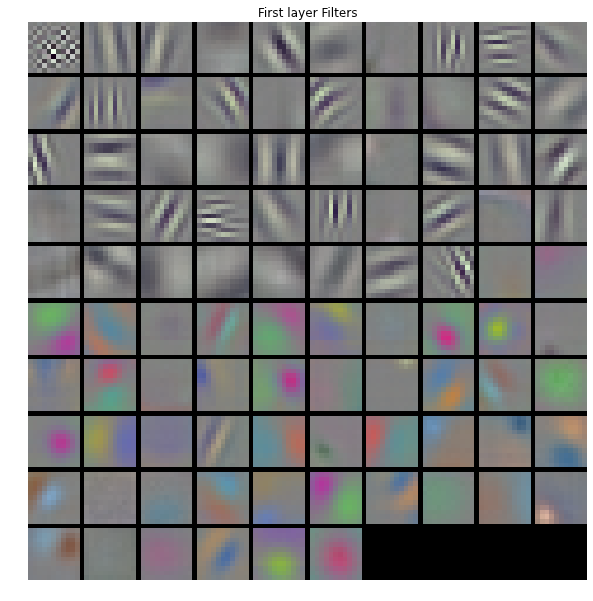}
\caption{The filters that appear in the first layer of the neural network once training is done.}
\label{fig:firstlayer}
\end{figure}

It can be noted that the trained filters from the first layer become Gabor filters, as visible from Figure \ref{fig:firstlayer}\footnote{Picture taken from http://nbviewer.ipython.org/github/UCB-ICSI-Vision-Group/decaf-release/blob/master/decaf/demos/notebooks/lena\_imagenet.ipynb for simplicity.}, with the subsequent layers providing higher-level representations of the features extracted from that first layer.

\subsection{Training}
The dataset chosen for this project is the Extended Cohn-Kanade Dataset (CK+) \cite{lucey2010extended}\footnote{Available at http://www.pitt.edu/\textasciitilde emotion/ck-spread.htm}. The dataset consists of 327 sequences acted out by 118 participants (the number of sequence per participants varies between 1 to 6), labelled by judges for the following emotions: anger, contempt, disgust, fear, happiness, sadness, and surprise.

The sequences can range from 10 to 30 frames, where the first frame shows the participant in a neutral state and the last frame (also called the peak frame) shows the participant in his or her most visible expression of the labelled emotion. Only the peak frames are used for training since no video stream is necessary for the system described in this paper.

As some sequences from this dataset are in grayscale and others are in color, every sequence is grayscaled before training\footnote{See discussion for a short justification}. Face detection is also applied to the images as a preprocessing step, where only the pixels within the rectangle detected by the Viola-Jones detector \cite{viola2001rapid} implemented in OpenCV \cite{opencv} are processed. The effect of applying or ignoring face detection to the images of the dataset is reported in the Results section.

\subsection{Classification}
To classify the features extracted from the still images, a Support Vector Machine model was chosen, as it is a commonly used classifier for this field of research \cite{khan2012human,lucey2010extended}. Two popular implementations, LIBLINEAR and LIBSVM \cite{REF08a,CC01a}, are packaged by the Python library scikit-learn \cite{scikit-learn} and were tested.

A few strategies exist for multiclassing SVMs \cite{hsu2002comparison}, and two of the most commonly used are compared later in this paper: ``one-versus-one" and ``one-versus-all". The ``one-versus-one" method is tested with multiple kernels and values of C, the soft margin parameter, but the ``one-versus-all" method is part of LIBLINEAR and is therefore limited to a linear kernel.

The dataset is also unbalanced, representing some emotions, such as surprise, more often than others. Scikit-learn offers the option of assigning class weights in relation to their frequency automatically, which is enabled here in order to account for the unbalanced data. 

To identify the best inter-participant performance, the scheme ``leave-one-participant-out" was used, producing 118 training and test sets so that the classifier would be tested on novel faces. This evaluation method maximizes the use of the data available (as there is only 327 sequences) and is consistent with the baseline experiments performed by Lucey et al. \cite{lucey2010extended}. The measure that is most commonly reported is the average of the accuracy across emotions, i.e. equal weights are given to the seven emotions even if some of them only have a few instances (``fear" is represented in 25 sequences, whereas there are 83 for ``surprise").

It should be noted that the performances for different values of the SVM parameters are shared in the Results section because no validation set was offered by Lucey et al. \cite{lucey2010extended} and the performances reported in some other papers, such as \cite{khan2012human} or \cite{zhao2007dynamic}, do not seem to use a validation set to select their parameters. This is done to prevent reporting a single best performance that would be biased.

\subsection{Live recognition}
To enable real-time emotion recognition during a video gaming session, a multithreaded application requiring a webcam was developed. The video game consists of using the arrow keys on the keyboard to avoid incoming debris on a 2D plane, losing health when a collision is detected. The rate of the the incoming debris is controlled by the facial expression of the player, decreasing when the player seems happy and increasing otherwise, forcing essentially the player to look happy in order to survive. It is hence called the ``Happiness game", and was inspired by a facial feedback effect reported in psychology, where smiling can accentuate a positive experience \cite{soussignan2002duchenne}.

In the main thread of the application, a video stream is captured from the player's webcam continuously with the OpenCV library \cite{opencv}. The player's face is first found using the Viola-Jones detector \cite{viola2001rapid} implemented in OpenCV. The library's implementation offers the use of different classifiers and parameters for the detector, and the performance of frontalface-alt2 was found to be reliable when used with a scale factor of 1.3 and a mininum number of neighbors of 3. The minimum size of a face was also set to a square of 150 pixels to reduce computational load and avoid detecting faces of other people standing far from the webcam. The frame is cut according to the location found by the detector and passed on to the secondary thread, where the features are extracted from the grayscaled frame and classified. With the processor tested, the AMD Phenom II X4 955\footnote{Slightly slow compared to currently available desktop processors}, 5 frames can be processed every second, which makes it viable for a real-time application.

The result of the classification is then passed to the main thread, which appends it to a list keeping the 5 most recent emotions detected. The most common emotion listed is then assigned to the player as his current emotion. This is done in order to keep the current emotion stable from single misclassified frames.

\begin{figure}[!t]
\centering
\includegraphics[width=\linewidth]{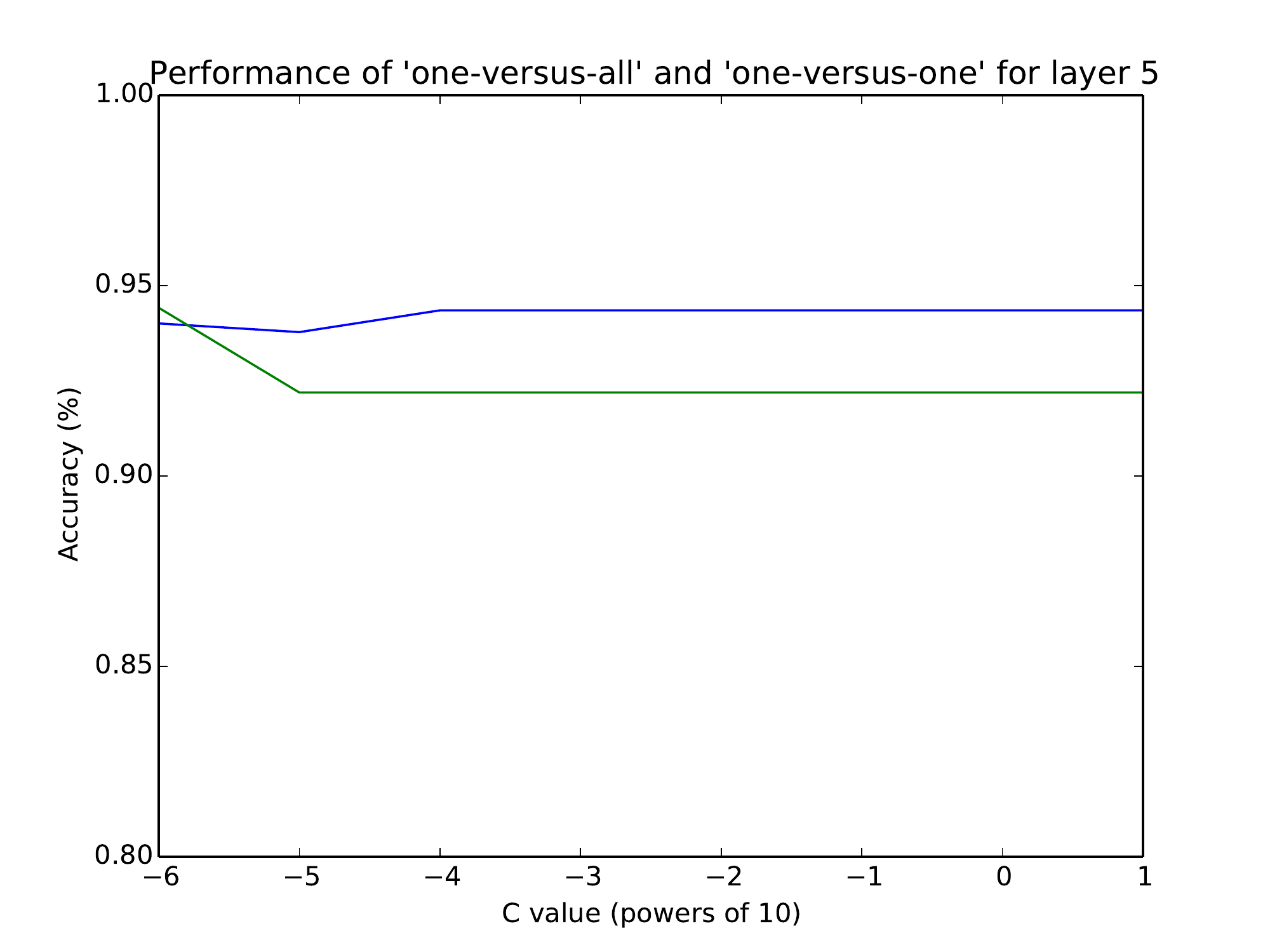}
\caption{A comparison of the multiclassing strategies for layer 5. The blue line is for the ``one-versus-all" method, and the green line is for the ``one-versus-one" method.}
\label{fig:5versusface}
\end{figure}

\section{Results}
The performance for multiple features and models are reported below for a ``leave-one-participant-out" scheme. The best performance overall was 94.4\% for a ``one-versus-one" SVM trained with a linear kernel, a C value of 1e-6, and with features taken from the fifth layer. Comparisons to other approaches will be made in the following Discussion section.

\begin{figure}[!t]
\centering
\includegraphics[width=\linewidth]{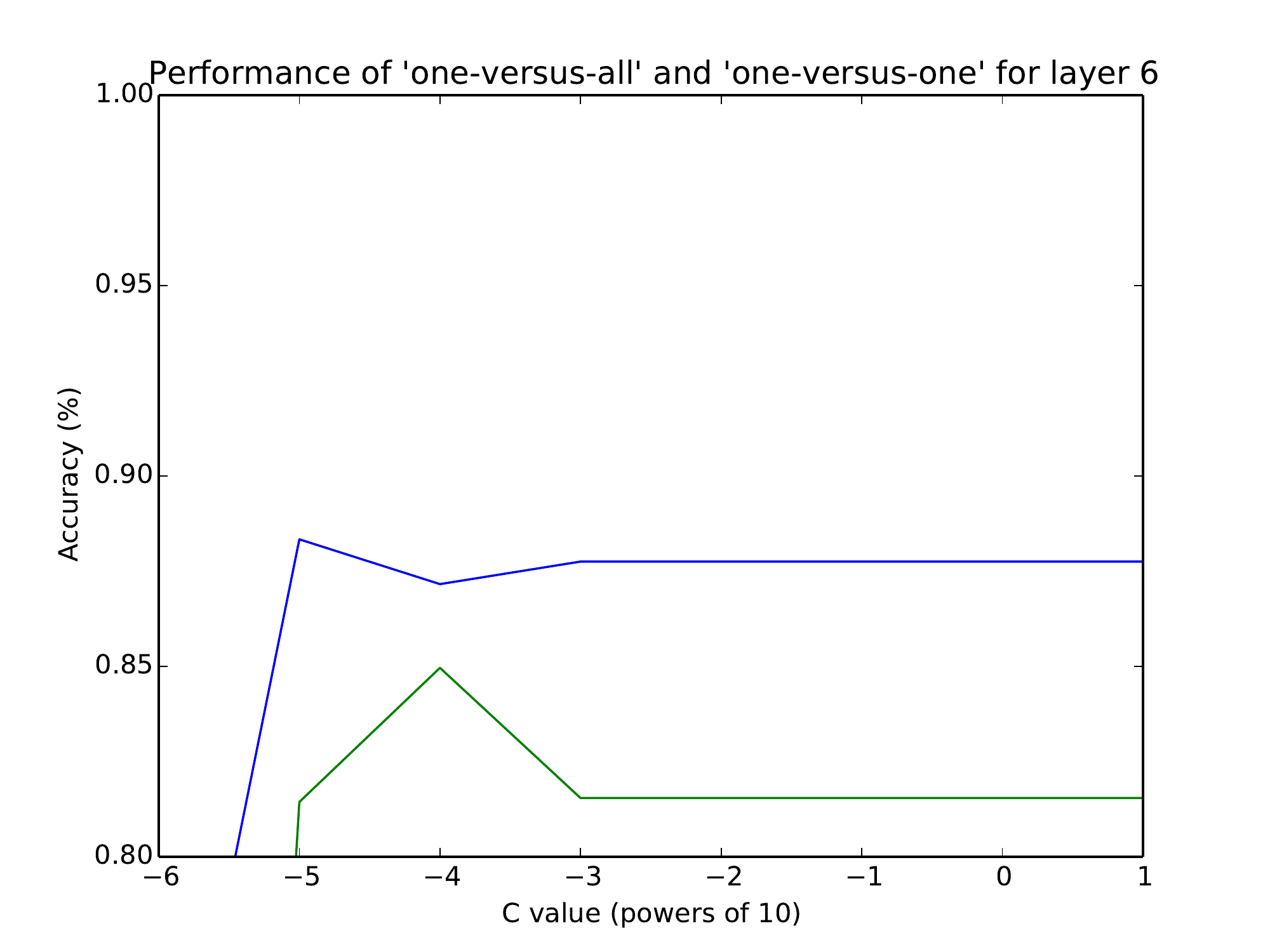}
\caption{A comparison of the multiclassing strategies for layer 6. The blue line is for the ``one-versus-all" method, and the green line is for the ``one-versus-one" method.}
\label{fig:6versusface}
\end{figure}

\begin{figure}[!t]
\centering
\includegraphics[width=\linewidth]{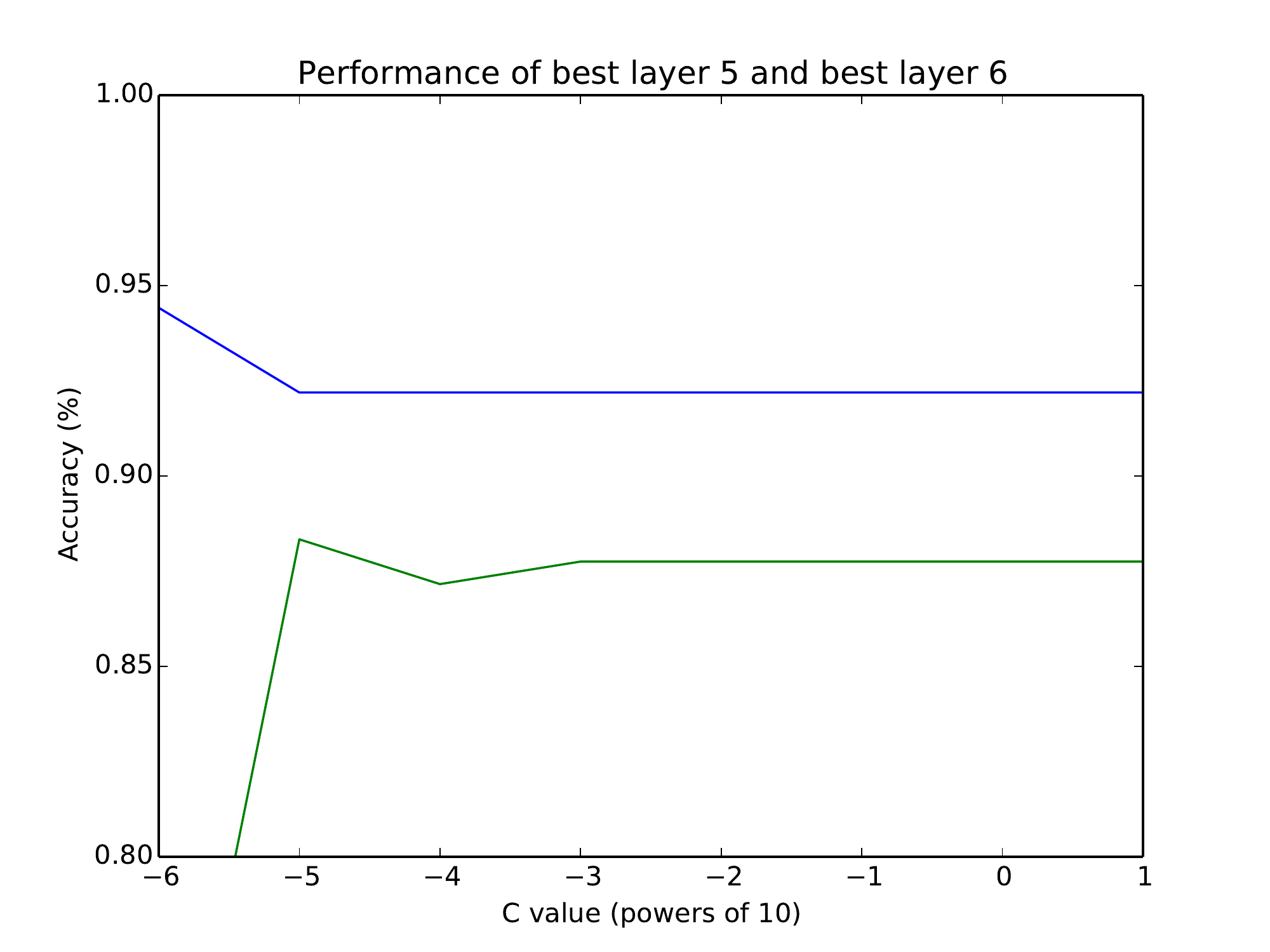}
\caption{A comparison of the layers given their best multiclassing strategy. The blue line is for the ``one-versus-all" method with features from the layer 5, and the green line is for the ``one-versus-all" method with features from the layer 6.}
\label{fig:56face}
\end{figure}

\subsection{Comparison of models with face detection}

Figure \ref{fig:5versusface} compares the performances with different values of C for the two different multiclassing strategies tested for features from the layer 5. The method ``one-versus-all" performs better for most tested values of C, and the effect is around 2\%. On the other hand, as seen in Figure \ref{fig:6versusface}, ``one-versus-all" outperforms the other method consistently. Both of these comparisons were done with a linear kernel.

Performance is however more affected by the layer from which the features are outputted, as Figure \ref{fig:56face} shows. The features from the fifth layer outperform those from the sixth layer consistently by at least 5\%.

In terms of the different kernels tested, radial-basis function and polynomial kernels were experimented with but offered lower performance (the highest seen was 90.7\% with a polynomial kernel of degree 2), were more sensitive to parameter selection, and experienced longer training times. Therefore, a more extensive evaluation of their performance is not reported here.

In order to assess the performance for individual emotions, Table 1 shows a confusion matrix for the best model found with the approach described above, where the true labels are on the vertical axis and the predicted labels are on the horizontal axis. The emotions (anger, contempt, disgust, fear, happiness, sadness, and surprise) are indicated by their first two letters. The accuracies are reported in percentage. The average accuracy across emotions is 94.4\%.

\begin{table}[!t]
\large
\renewcommand{\arraystretch}{1.3}
\caption{Confusion matrix for the CK+ dataset with face detection}
\centering
\begin{tabular}{ l | c  c  c c c c c }
  - & An & Co & Di & Fe & Ha & Sa & Su \\ \hline  
 An&\textbf{91.1}&    2.2&    0. &    0. &    0. &    6.7&    0. \\
 Co&   0. &  \textbf{100.} &    0. &    0. &    0. &    0. &    0. \\
 Di&  5.1&    0. &   \textbf{93.2}&    0. &    1.7&    0. &    0. \\
 Fe&   0. &    0. &    0. &  \textbf{100.} &    0. &    0. &    0. \\
 Ha&   0. &    1.4&    0. &    2.9&   \textbf{95.7}&    0. &    0. \\
 Sa&  17.9&    0. &    0. &    0. &    0. &   \textbf{82.1}&    0. \\
 Su&   0. &    1.2&    0. &    0. &    0. &    0. &   \textbf{98.8}\\
\end{tabular}
\end{table}

\subsection{Comparison of models without face detection}

Removing face detection, i.e. letting the whole 640x480 image be processed by the convolutional network, decreases the performance of the models considerably.  An accuracy of 77.3\% was found with a ``one-versus-one" SVM trained with a linear kernel and C value of 1e-4, with features taken from the fifth layer. Figures (5 to 7) and a confusion matrix (Table 2) for the best model without face detection are shown without further description.

\begin{table}[!t]
\large
\renewcommand{\arraystretch}{1.3}
\caption{Confusion matrix for the CK+ dataset without face detection}
\centering
\begin{tabular}{ l | c  c  c c c c c }
  - & An & Co & Di & Fe & Ha & Sa & Su \\ \hline  
An& \textbf{66.7}&   0. &  15.6&   2.2&   0. &  15.6&   0.    \\ 
Co& 0. &  \textbf{77.8}&   0. &   0. &  11.1&   5.6&   5.6 \\ 
Di& 5.1&   0. &  \textbf{89.8}&   0. &   1.7&   0. &   3.4   \\ 
Fe& 4. &   0. &   0. &  \textbf{52.} &  24. &  12. &   8. \\ 
Ha& 1.4&   0. &   1.4&   1.4&  \textbf{95.7}&   0. &   0.   \\ 
Sa& 14.3&   7.1&  10.7&   0. &   0. &  \textbf{60.7}&   7.1   \\ 
Su& 0. &   1.2&   0. &   0. &   0. &   0. &  \textbf{98.8} \\
\end{tabular}
\end{table}

\begin{figure}[!t]
\centering
\includegraphics[width=\linewidth]{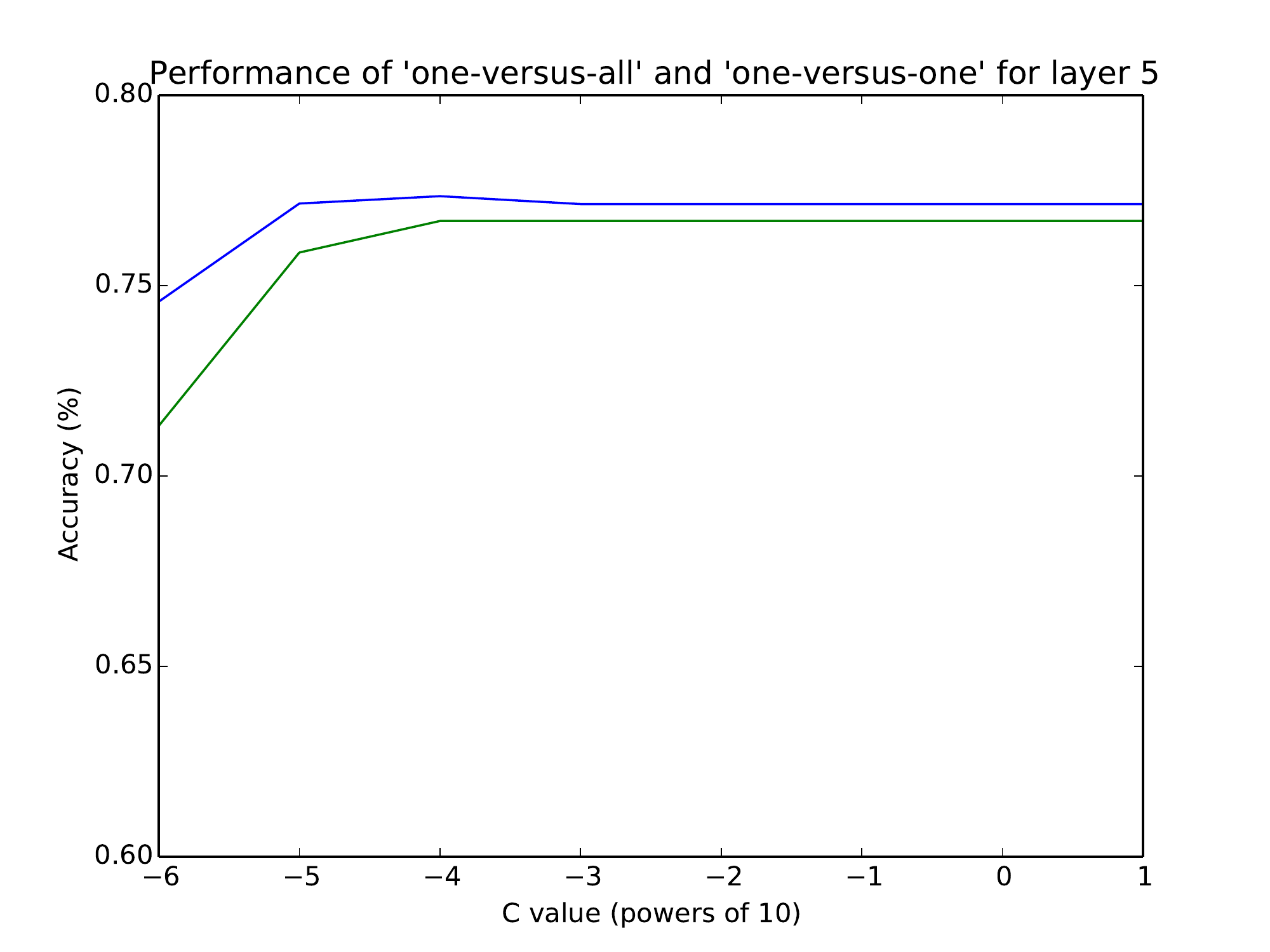}
\caption{A comparison of the multiclassing strategies for layer 5. The blue line is for the ``one-versus-all" method, and the green line is for the ``one-versus-one" method.}
\label{fig:5versus}
\end{figure}

\begin{figure}[!t]
\centering
\includegraphics[width=\linewidth]{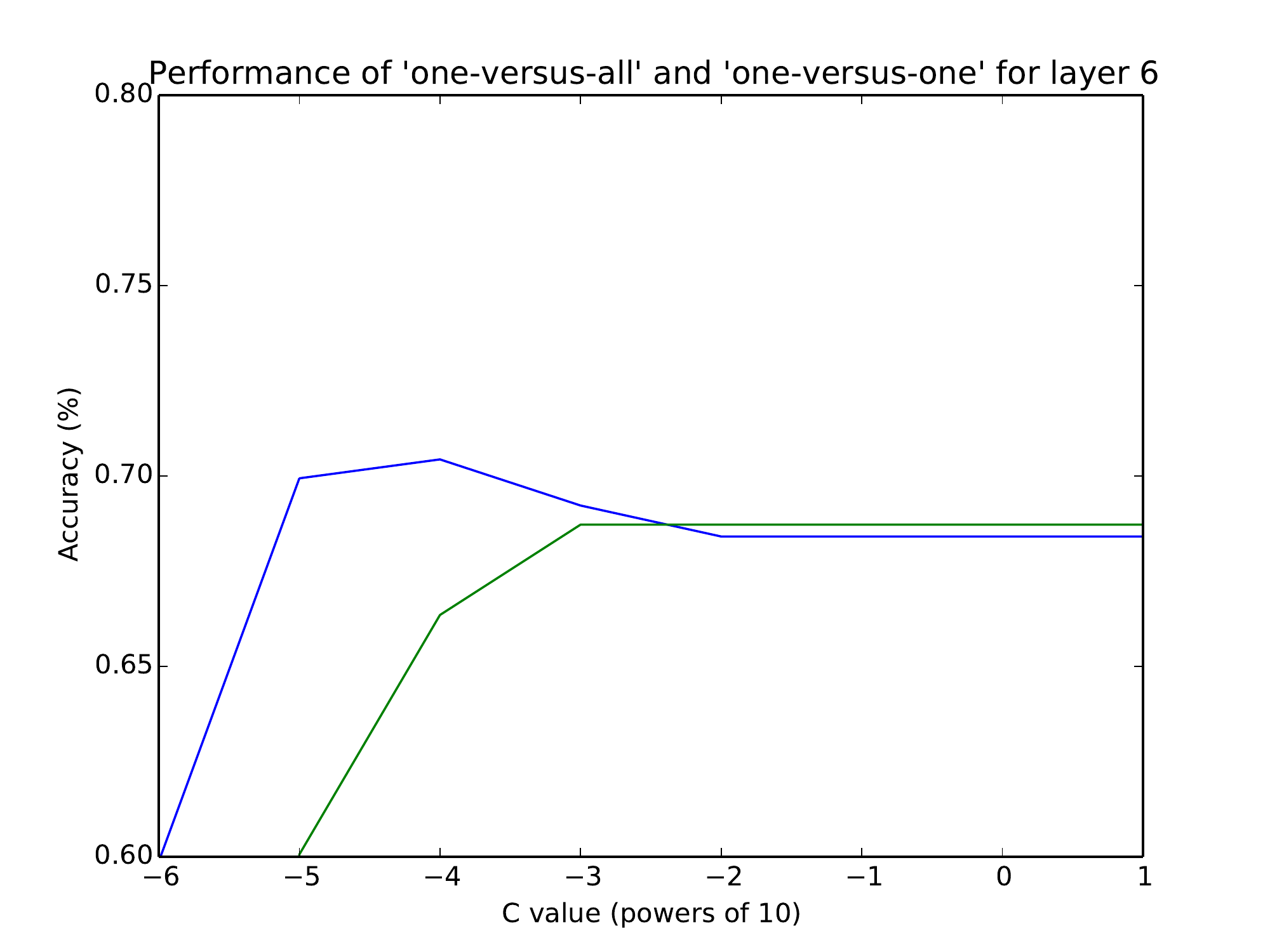}
\caption{A comparison of the multiclassing strategies for layer 6. The blue line is for the ``one-versus-all" method, and the green line is for the ``one-versus-one" method.}
\label{fig:6versus}
\end{figure}

\begin{figure}[!t]
\centering
\includegraphics[width=\linewidth]{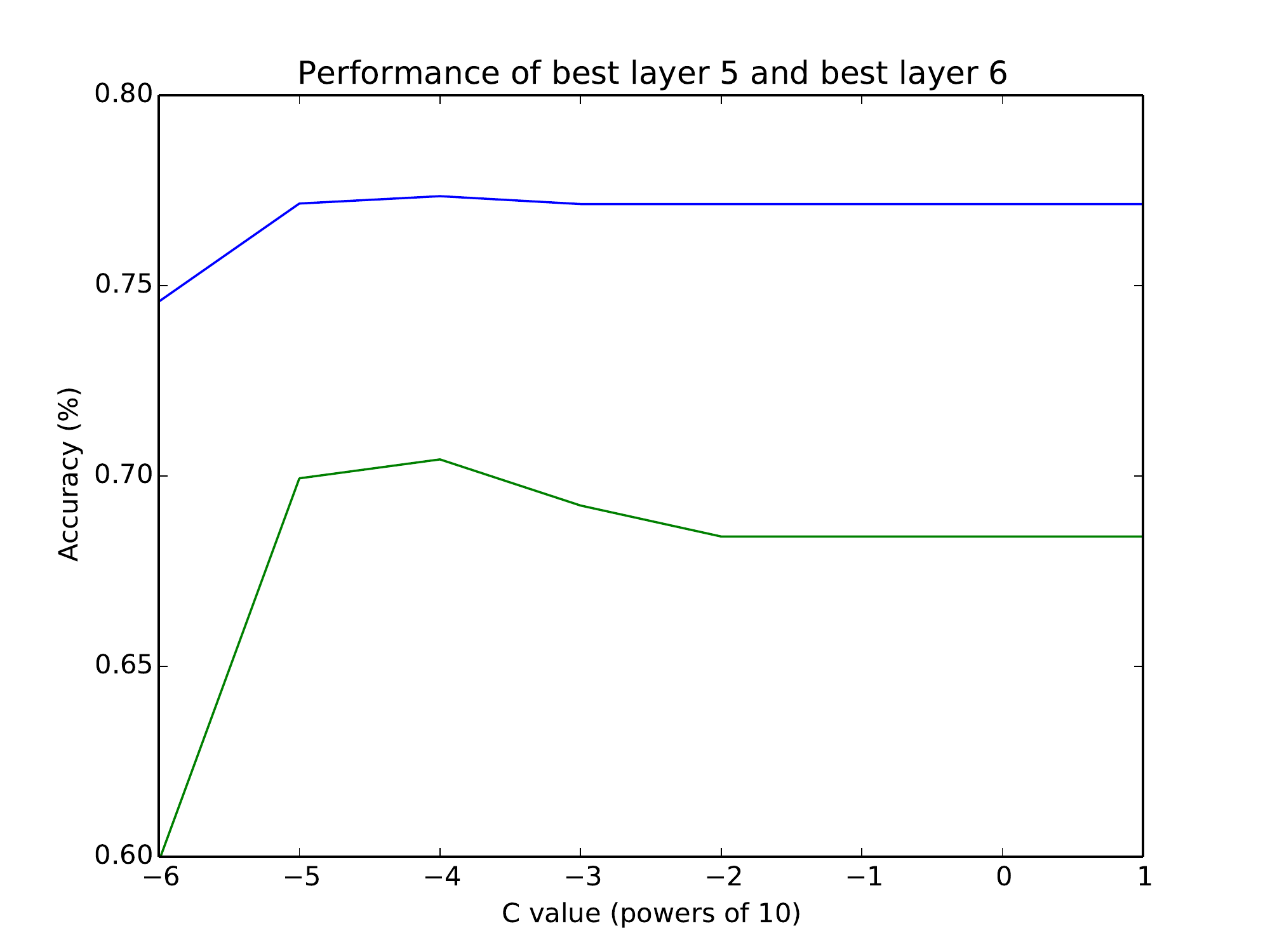}
\caption{A comparison of the layers given their best multiclassing strategy. The blue line is for the ``one-versus-all" method with features from the layer 5, and the green line is for the ``one-versus-all" method with features from the layer 6.}
\label{fig:56}
\end{figure}

\subsection{Qualitative assessment of the real-time gaming experience}
The affective feedback game was solely tested by myself, the writer of this paper, and therefore this section is highly subject to bias until proper experiments can be performed with participants. It can still be pointed out that the application seemed to accurately predict my facial expressions most of the time, and that the experience seemed uniform despite changes in lighting conditions or distances from the camera.

\begin{figure}[!t]
\centering
\includegraphics[width=\linewidth]{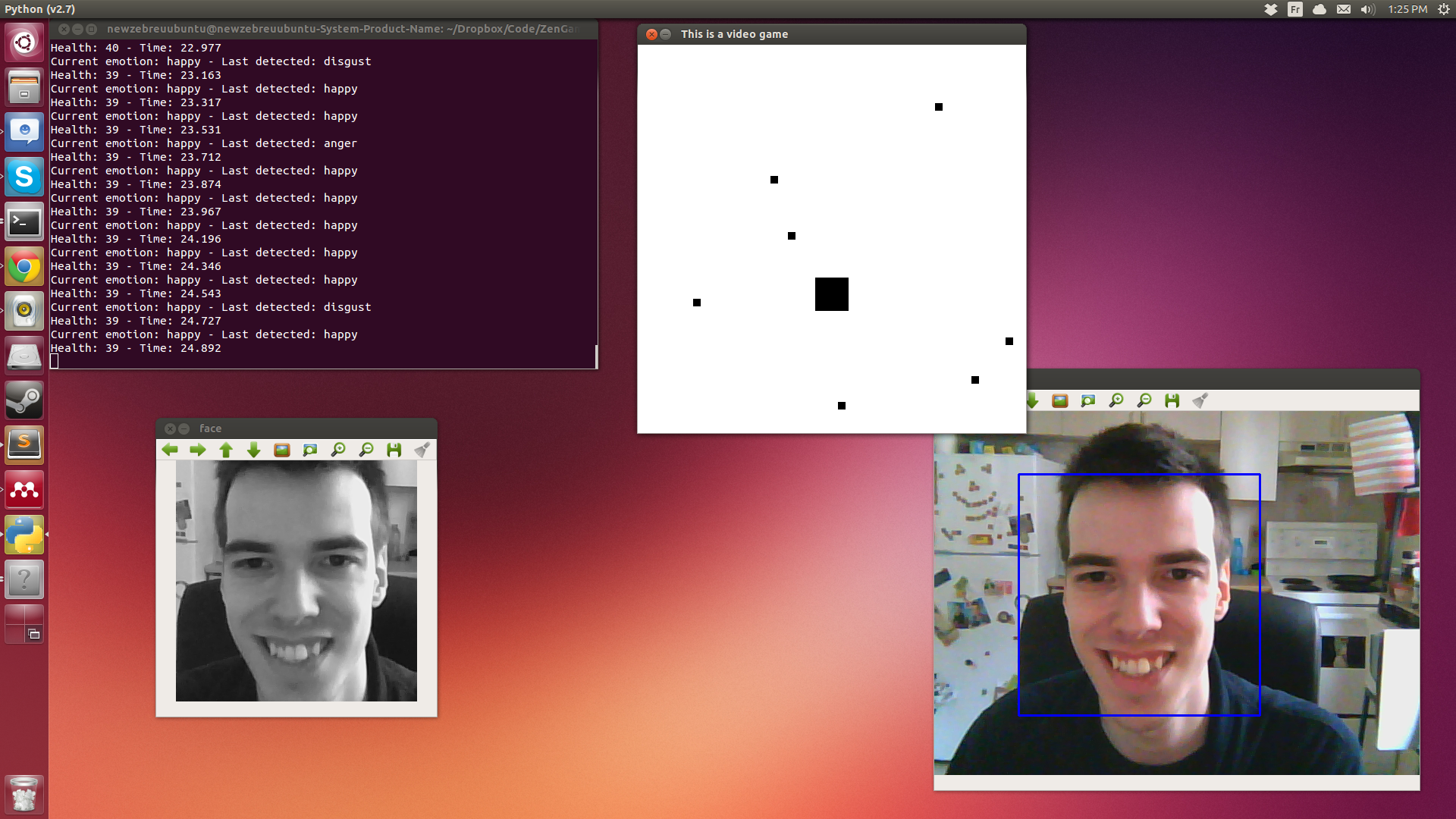}
\caption{A screenshot of a gaming session. The number of debris is at its lowest since ``happiness" is detected, as shown in the terminal window to the top left if you zoom in. We can see that a single frame was detected as if I was ``disgusted", but the current emotion determined from the 5-snapshots window is stable nonetheless.}
\label{fig:happy}
\end{figure}

\begin{figure}[!t]
\centering
\includegraphics[width=\linewidth]{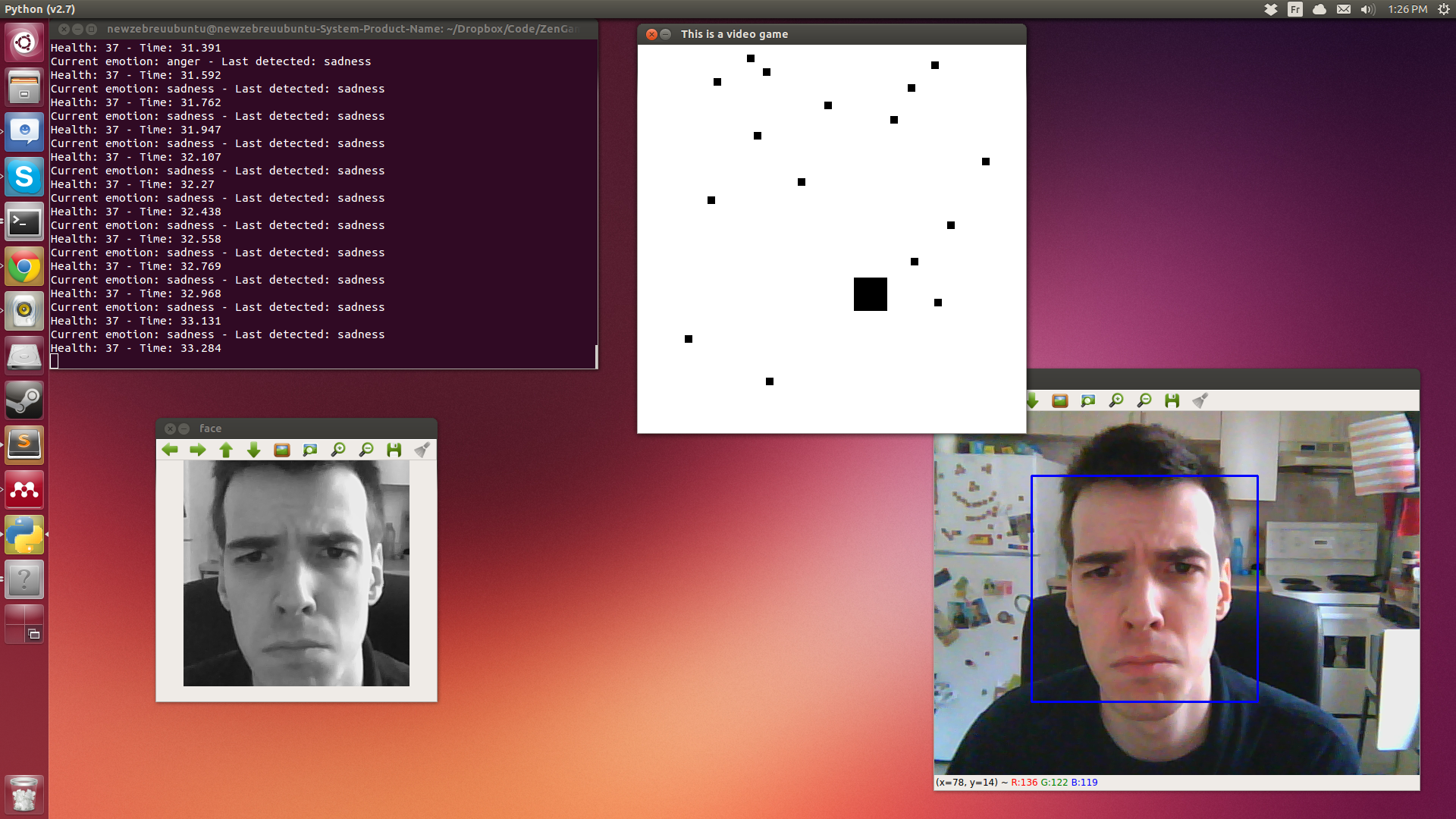}
\caption{A screenshot of the same gaming session as in the previous figure, but where I show ``sadness" and its detection triggers the game into creating more debris.}
\label{fig:sad}
\end{figure}

\section{Discussion}
The best model found had a performance of 94.4\%, 11.1\% above the baseline reported in \cite{lucey2010extended}, and 4.7\% below the state-of-the-art model reported in \cite{jeni2013continuous}. Other recent specialized approaches perform similarly \cite{khan2012human}. 

These other approaches seem more robust in terms of accuracy per emotion, where none of them is misclassified considerably more often than the others, which is not the case for the approach tested in the present study with regards to ``sadness" with 82.1\%, and ``anger" to a lesser degree.

The models developed for this paper that do not use face detection on the images from the CK+ dataset suffer from a much wider imbalance across emotions, which decreases the overall accuracy by 17.1\% compared to the best model tested. Face detection therefore seems to be a very useful preprocessing step even with centered and uniform pictures. The main weaknesses of the model without face detection seem to be the emotions ``fear" and ``sadness", at 52\% and 60.7\% respectively. The baseline system \cite{lucey2010extended} also had a lower recognition rate for these two emotions. A partial explanation for the lower performances might be related to the number of sequences which present these emotions, as those are among the emotions with the fewest instances. 

The features outputted from the fifth and sixth layers of the network were tested, but the seventh layer was ignored for two reasons. First, the performance degraded from the fifth to the sixth, so it seemed unlikely that the seventh layer would produce better features. Second, the sixth and seventh layers are possibly optimized for the object classification task on which they were trained, as they are fully-connected layers rather than convolutional layers. It seemed therefore more likely that the output from the fifth layer encoded most of the image information unrelated specifically to object classification, and that the next two layers transformed it to improve performance on the ImageNet Challenge. The fifth layer was also shown to perform better on some different tasks, such as determining the aesthetic rating of a picture \cite{karayev2013recognizing}.

Most of the sequences of the CK+ dataset were recorded in grayscale, and to uniformize the data, all images were therefore grayscaled. Before this change was done, the performance observed was slightly lower. However, the convolutional network was trained on color images, and it might benefit from a dataset of color images. 

Linear kernels were shown to perform better at a lower cost in terms of computational time compared to polynomial or radial basis function kernels. The radial basis function seemed to perform considerably worse than the others, but it might be due to the limited grid-search (of gamma and C values) done. Such differences in performance can be explained by the high dimensionality of the features, especially considering the 9216 dimensions of the output from the fifth layer. Non-linear kernels generally increase the number of dimensions in order to find a space where the dataset is accurately separable, and are therefore most useful for features of low dimension, such as the ones used by Khan et al. for their approach, namely 168-dimensional features \cite{khan2012human}. Given the already high dimensionality of the convolutional features, linear kernels therefore seem to be the best option here.

\subsection{Limitations and future work}
The current approach processes only the apex frame of the sequences, comparatively to most other approaches where a video stream is processed. Using the apex frame accentuates the distinction between emotions, avoiding frames where the expressions are more subtle, but such an effect might be balanced by limiting the system to a single frame rather than the motion information that could be extracted from the sequences. It would be possible to use such motion information in a later version of the system in multiple ways. One possible method would be similar to the one employed for the real-time application: aggregating the predictions from each frame, and then assigning the most commonly predicted label to the sequence. Another method would be to stack the features sequentially and let the classifier handle very high-dimensionality data points.

The CK+ dataset doesn't seem to offer information on whether some of the participants were wearing glasses. While testing the real-time application with glasses, when a model trained without face detection was used, it was observed that ``disgust" was often detected instead of other emotions until the glasses were removed. This is not surprising as facial occlusions are generally hard to handle \cite{fasel2003automatic}. On the other hand, when tested with the best model found (integrating face detection), the glasses were not an issue anymore. Given this observation, it would be interesting to test the model on participants that present facial occlusions such as headwear, glasses, or facial hair.

\section{Conclusion}
The present study investigated the application of pre-trained features from a deep convolutional neural network to the task of emotion recognition. The network was initially trained on object recognition, and no further training of the network took place on the emotion recognition task. With the simple preprocessing step of detecting faces using the Viola-Jones detector, a SVM using these features classifies seven emotions (as labelled in the CK+ dataset) with an accuracy of 94.4\%.

The results presented in this paper support the use of these features in a generic visual system, offering very good performance across many tasks \cite{donahue2013decaf, karayev2013recognizing} with no retraining or extensive preprocessing required.

\ifCLASSOPTIONcaptionsoff
  \newpage
\fi

\bibliographystyle{ieeetr}
\bibliography{bibliography.bib}

\vfill
\begin{IEEEbiographynophoto}{S\'ebastien Ouellet}
is a graduate student at Universit\'e de Montr\'eal, mostly interested in artificial intelligence. His current line of research (adaptive video games) is influenced by his previous degree, a Bachelor of Cognitive Science from Carleton University.
\end{IEEEbiographynophoto}

\end{document}